\documentclass[letterpaper, 10 pt, conference]{ieeeconf}
\IEEEoverridecommandlockouts
\usepackage{times}
\usepackage{graphicx}
\usepackage{amsmath}
\usepackage{amssymb}

\usepackage{amsthm}
\usepackage{color}
\usepackage{multicol}
\usepackage{multirow}
\usepackage{subcaption}
\usepackage{array}

\DeclareMathOperator{\atan2}{atan2}
 
\DeclareMathOperator{\sgn}{sgn}

\newcolumntype{C}[1]{>{\centering\arraybackslash}p{#1}}

\title{\LARGE \bf
LaserFlow: Efficient and Probabilistic Object Detection and\\Motion Forecasting
}

\author{Gregory P. Meyer$^{1}$, Jake Charland$^{1}$, Shreyash Pandey$^{1}$, Ankit Laddha$^{1}$, Shivam Gautam$^{1}$,\\
        Carlos Vallespi-Gonzalez$^{1}$, and Carl K. Wellington$^{1}$%
\thanks{$^{1}$Uber Advanced Technologies Group}%
\thanks{Correspondence to {\tt\small gmeyer@uber.com}}%
}

\begin{document}

\maketitle
\thispagestyle{empty}
\pagestyle{empty}

\begin{abstract}
In this work, we present LaserFlow, an efficient method for 3D object detection and motion forecasting from LiDAR.
Unlike the previous work, our approach utilizes the native range view representation of the LiDAR, which enables our method to operate at the full range of the sensor in real-time without voxelization or compression of the data.
We propose a new multi-sweep fusion architecture, which extracts and merges temporal features directly from the range images.
Furthermore, we propose a novel technique for learning a probability distribution over future trajectories inspired by curriculum learning.
We evaluate LaserFlow on two autonomous driving datasets and demonstrate competitive results when compared to the existing state-of-the-art methods.
\end{abstract}

\section{Introduction}
\label{sec:intro}

\begin{figure*}[t]
    \centering
    \begin{subfigure}{0.48\textwidth}
        \centering
        \includegraphics[height=2.5in]{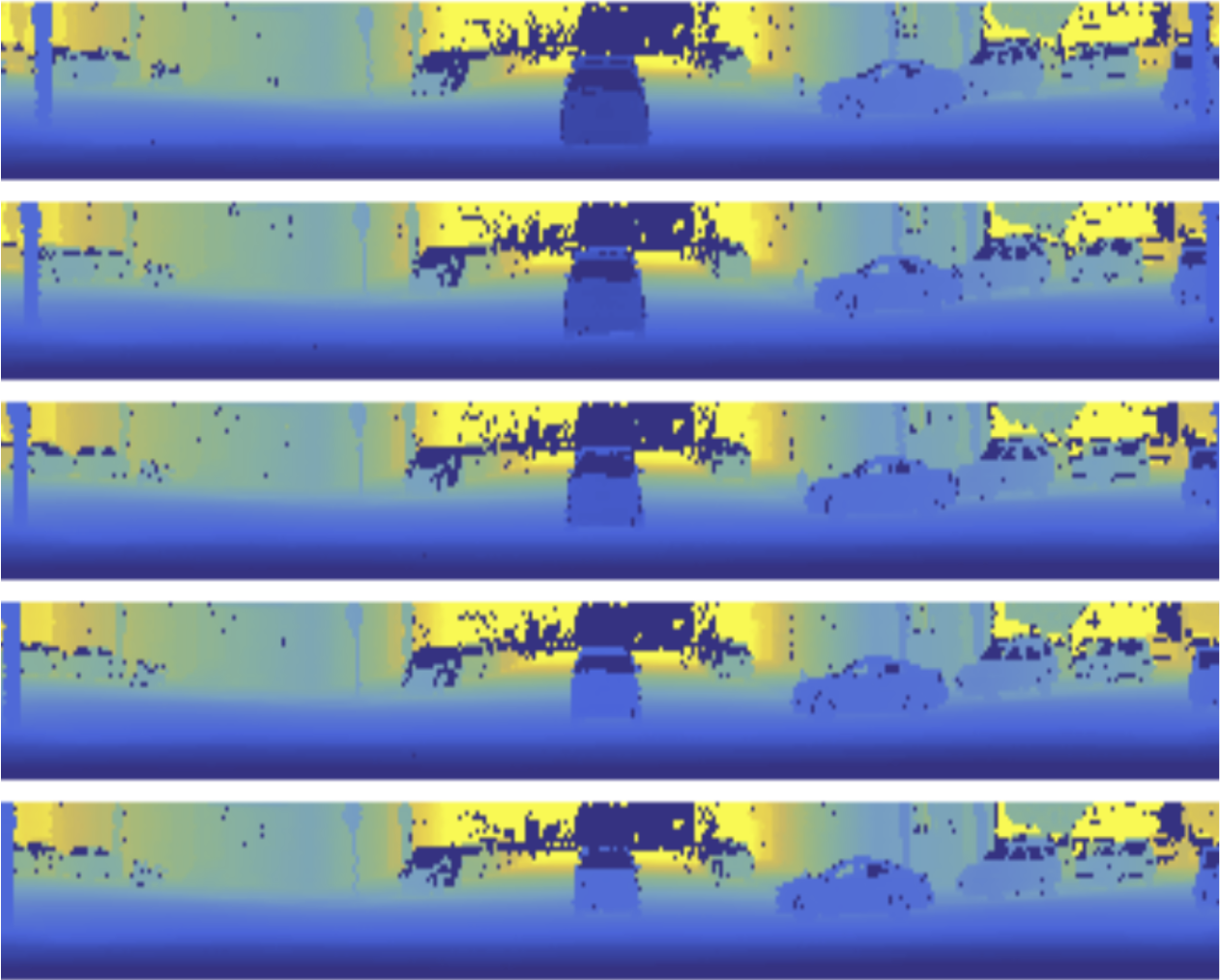}
        \caption{Input sequence of LiDAR range data}
        \label{fig:summary:input}
    \end{subfigure}
    \begin{subfigure}{0.48\textwidth}
        \centering
        \includegraphics[height=2.5in]{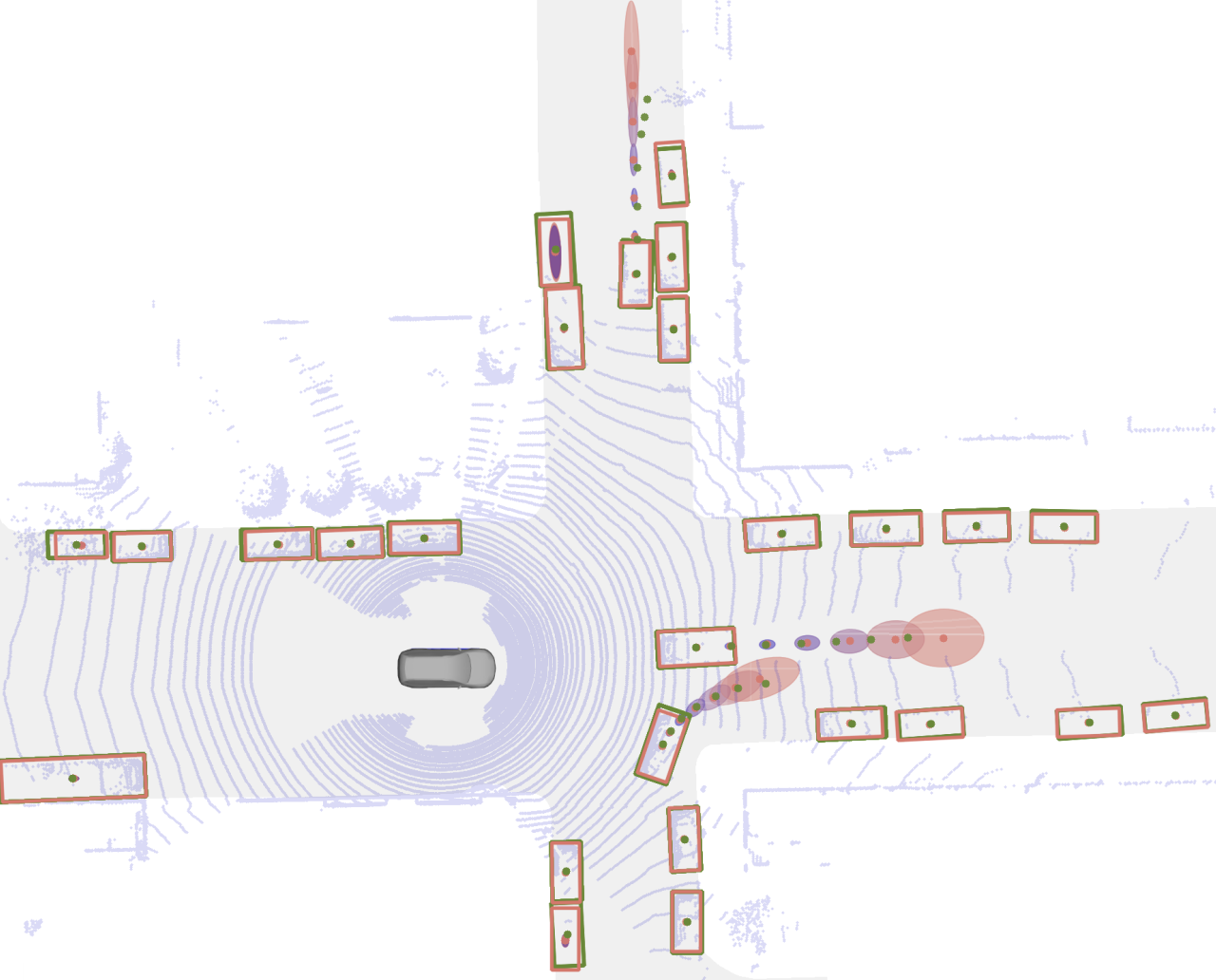}
        \caption{Output predictions with uncertainty}
        \label{fig:summary:output}
    \end{subfigure}
    \caption{End-to-end learning of motion predictions from sensor data takes as input the native range view (RV) measurements produced by LiDAR sensors and generates output in the bird's eye view (BEV) as needed by downstream components. Unlike prior work which starts by transforming the range data into 3D points and then processing the resulting sparse data in the BEV, we operate directly in the compact native RV of the sensor.}
    \label{fig:summary}
    \vspace{-1em}
\end{figure*}

The detection and motion forecasting of objects represent critical capabilities for a self-driving system.
Many algorithms proposed for forecasting object motion take a sequence of detections as their input, but more recently there has been a focus on end-to-end approaches that generate predictions of future motion directly from a temporal sequence of the sensor data~\cite{spagnn,intentnet,faf,nmp}.
LiDAR range sensors are commonly used for this task, and Figure~\ref{fig:summary} shows an example sequence of the raw input LiDAR sweeps in their native range view (RV) format.
Also shown in Figure~\ref{fig:summary} is the desired output, which includes object detections along with probability distributions over their predicted future motion in the bird's eye view (BEV).
Given that LiDAR sensors inherently produce a set of range measurements, but the desired output motion forecasts are in the BEV, an important design consideration for end-to-end approaches is how to best process RV measurements to produce BEV output.

Prior work starts by transforming the raw range measurements from the sensor into a 3D point cloud in a global coordinate frame, and then this sequence of aligned 3D point clouds is used to predict the motion of actors in the scene.
The 3D points are generally accumulated in a grid of cells or voxels to allow convolutions to be performed in the BEV which matches the desired output representation.

BEV approaches have limitations in runtime efficiency, which is important for a real-time system that needs to react quickly to a changing environment.
As shown in Figure~\ref{fig:summary:output}, the projected 3D points are sparse, and performing feature extraction in the BEV can require defining a limited region of interest (RoI) for processing~\cite{spagnn,intentnet,faf,nmp}.
Furthermore, BEV approaches that process the 3D points into cells have a trade-off between performance and runtime that is dependent on the cell size~\cite{pointpillars}.

Alternatively, extracting features in the RV, where the data is naturally dense, supports efficiently operating out to the maximum range of the sensor.
The sensor data in the native RV also captures important context about which parts of the scene are visible by the sensor and which parts are occluded by other objects.
This occlusion information is lost when the data is transformed into a 3D point cloud.
Finally, operating on the raw measurements of the sensor removes the complexity and trade-offs of artificial abstractions like cells and voxels, resulting in an approach that can operate efficiently at all ranges while still performing well on smaller objects~\cite{lasernet, lasernet++}.

This work extends the RV based probabilistic object detection introduced in~\cite{lasernet} to the problem of motion forecasting.
A key challenge to using the RV for motion forecasting is how to combine a sequence of multiple LiDAR sweeps that were captured at different sensor locations due to vehicle ego-motion.
Projecting this temporal data into a common RV frame to extract multi-sweep features can result in self-occlusions and gaps in the data due to the change in perspective, and our proposed method is designed to overcome these challenges.

In this paper, we propose a highly efficient method for probabilistic object detection and motion forecasting using LiDAR.
We achieve competitive performance to state-of-the-art methods on the task of motion forecasting with an efficient runtime that supports operating over the full range of the sensor. 
This is achieved by extracting features on the raw range view measurements produced natively by the LiDAR sensor and then using a novel method of fusing multiple sweeps.
Finally, using an approach inspired by curriculum learning, we present a novel approach for learning a distribution over future trajectories that initially emphasizes earlier predictions before learning longer horizons.

\section{Related Work}
\label{sec:related-work}

\subsection{3D Object Detection from LiDAR}
Over the past few years, a wealth of approaches~\cite{pointpillars,lasernet,lasernet++,zhouVoxelNetEndtoEndLearning2018,second,yangPIXORRealtime3D2018, pointrcnn,std,mv3d,EndtoEndMF} have been proposed to solve 3D object detection using LiDAR.
The methods typically vary in terms of the representation of the LiDAR data.
VoxelNet~\cite{zhouVoxelNetEndtoEndLearning2018} and SECOND~\cite{second} represent the world as a 3D grid and employ 3D convolutions.
PIXOR~\cite{yangPIXORRealtime3D2018} and PointPillars~\cite{pointpillars} demonstrated that a 2D grid and 2D convolutions could be used as an alternative to high-cost 3D convolutions. 
However, representing the world as a 2D or 3D grid is still inefficient due to the sparsity of the LiDAR, especially at long range, resulting in a significant amount of empty cells.
Furthermore, due to voxelization, these representations lose fine-grain geometrical details, which make the detection of small objects difficult.
As a result, PointRCNN~\cite{pointrcnn} and STD~\cite{std} proposed to use a point cloud representation.
However, by treating the LiDAR data as an unordered set of points, these methods lose information about how the scene was captured, which make it challenging to reason about occlusions.
Additionally, these methods are often slower due to the unstructured nature of the representation.
LaserNet~\cite{lasernet} proposed to use a spherical representation, referred to as the range view, where each pixel represents a LiDAR point.
The range view is the native representation of the LiDAR; therefore, it is compact and retains the fine-grain details.
In this work, we extend LaserNet~\cite{lasernet} to predict the motion of objects.

\subsection{Motion Forecasting from Detections}
Motion forecasting requires a history of observations.
Several previous methods \cite{desire, sociallstm, convolutionalsocialpooling, carnet, socialattention, sophie, tensorfusion, dp, rulesofroad, chauffeurnet, multipath} have used a sequence of past detections to predict future motion.
These methods can be broadly divided into two groups: approaches that use a feed-forward convolutional neural network (CNN) and methods that encode historical information using a recurrent neural network (RNN).
Methods that leverage RNNs \cite{desire,sociallstm, convolutionalsocialpooling,carnet,socialattention,sophie,mfp} often model the motion of each object independently and employ separate models for different interactions, such as actor to actor~\cite{sociallstm, convolutionalsocialpooling,socialattention}, actor to scene~\cite{carnet}, or both~\cite{desire,sophie}.
CNN based methods \cite{dp, rulesofroad, chauffeurnet} capture both object motion and interactions in a sequence of images and use a feed-forward network to predict the future.
For each actor, they render a set of images containing the position of the actor and other nearby objects across time.
Afterwards, these images are passed to a CNN which forecasts the motion for the object of interest.
Recently, \cite{tensorfusion,multipath} have been proposed which predict trajectories for all the actors together.
These approaches typically require some amount of computation to be run for every actor in the scene, making it difficult to scale to dense urban environments. 
Furthermore, all of these methods disregard the rich information provided by the sensor. 

\subsection{Motion Forecasting from LiDAR}
To our knowledge, the first method to attempt joint 3D object detection and motion forecasting from LiDAR was FaF~\cite{faf}.
Their approach used a 3D grid to represent a sequence of LiDAR point clouds and predicted a trajectory one second in length for all vehicles within a region of interest.
IntentNet~\cite{intentnet} extended FaF to predict the intent of each actor, e.g. keep straight, turn right,  merge left, etc., and increased the length of the trajectory to $3$ seconds.
Their method was further improved by NMP~\cite{nmp} which added the motion planner of the self-driving vehicle as an additional source of supervision to the model.
Most recently, SpAGNN~\cite{spagnn} introduced a graph-based interaction model and uncertainty estimates to IntentNet.
All of the previous work that performs joint object detection and motion forecasting uses a 3D or bird's eye view representation of the LiDAR.
In contrast, we propose to use the native range view representation of the LiDAR.

\section{Proposed Method}
\label{sec:method}

In the following sections, we will describe our approach to multi-sweep fusion (Section \ref{sec:multisweep}), our proposed method to predicting a probability distribution of trajectories (Section \ref{sec:predictions}), and our end-to-end training procedure (Section \ref{sec:training}).

\subsection{Multi-Sweep Fusion}
\label{sec:multisweep}

As the LiDAR spins, it produces an image using a set of range sensing lasers.
For each measurement, the sensor returns the range $r$ and reflectance $e$ of the observed surface.
In addition, it provides the azimuth angle of the sensor $\theta$ and elevation angle of the corresponding laser $\varphi$ from which the 3D position $\boldsymbol{x}$ of the surface can be calculated \cite{lasernet++}.
A complete revolution of the LiDAR is colloquially referred to as a sweep.
Multiple sweeps of the LiDAR can be utilized to capture the motion of objects across time.

Unlike the previous work \cite{spagnn,intentnet,faf,nmp}, our proposed method uses the native range view (RV) representation of the LiDAR data instead of the bird's eye view (BEV).
Using the RV has advantages in terms of efficiency and small object detection \cite{lasernet,lasernet++}; however, the fusion of multiple LiDAR sweeps becomes non-trivial.
In order to fuse measurements from multiple sweeps, the 3D points are transformed into a global coordinate frame that accounts for the ego-motion of the self-driving vehicle.
Afterwards, the points can be projected into a shared image frame where multi-sweep features are extracted using a convolutional neural network (CNN).
With the BEV, an orthogonal projection is used; therefore, the difference between the original and shared image frames amounts to a 2D rotation and translation.
For the RV, a spherical projection is used; therefore, transforming from the original to a shared image frame may cause a change in perspective.
As a result, shadows or holes from unobserved or partially observed surfaces and data loss due to self-occlusions may occur.

To overcome these issues, we propose a novel multi-sweep fusion architecture.
With the proposed architecture, features are extracted independently from each of the LiDAR sweeps in their original view.
Afterwards, the features from each sweep are passed through a feature transformer which learns to transform the features from the original coordinate frame to the global coordinate frame.
The previous sweeps are then warped into the current sweep's image frame and concatenated; at which point, the multi-sweep feature maps can be input into a backbone network to detect objects and estimate their motion.

\subsubsection{Input Features}
The input to our multi-sweep architecture is a set of images, $\{\boldsymbol{I}_S, \boldsymbol{I}_{S+1}, \ldots, \boldsymbol{I}_0\}$, corresponding to a sequence of sweeps.
The images are constructed by mapping LiDAR measurements, also referred to as points, to a column based on the azimuth angle of the sensor at the time the measurement was captured and to a row based on the elevation angle of the laser that produced the measurement.
For each pixel in the image, we generate a set of channels corresponding to the measurement's range, reflectance, and a flag indicating whether the measurement is valid.
Furthermore, if a high-definition map is available, we provide the LiDAR point's height above the ground and a flag indicating whether the point is on or above a road surface.

\subsubsection{Ego-Motion Features}
Since the self-driving vehicle moves over time, we provide the ego-motion as a feature.
Let ${\boldsymbol{P}_s \in \mathbb{SE}(3)}$ represent the pose of the LiDAR at the start of the $s$-th sweep.
The motion of the ego-vehicle from sweep $s$ to the current sweep can be computed as $\boldsymbol{\Delta}_s = \boldsymbol{P}_0 \boldsymbol{P}_s^{-1} [0,0,0,1]^T$.
Our proposed method utilizes a fully convolutional neural network; therefore, we rotate the ego-motion features based on the position within the image, i.e. the azimuth angle $\theta$,
\begin{equation}
    \boldsymbol{\delta} = \boldsymbol{R}^T_\theta [\Delta_s^x, \Delta_s^y]^T
    \label{eq:egomotion}
\end{equation}
where $\boldsymbol{R}_\theta$ is the rotation matrix parameterized by $\theta$, and $\Delta_s^x$ and $\Delta_s^y$ are the $x$ and $y$ component of $\boldsymbol{\Delta}_s$, respectively.
The importance of rotating the ego-motion feature is shown in our ablation study.

\subsubsection{Feature Warping}
To transform the features between sweeps, we need to define a mapping between all pixels in one image to another.
This process is referred to as feature warping.
Each valid pixel corresponds to a LiDAR point $\boldsymbol{x}$, and each image has a corresponding pose $\boldsymbol{P}_s$.
To map from a previous sweep to the current sweep, we first transform the point into the current sweep's coordinate frame, ${\boldsymbol{x}' = \boldsymbol{P}_0 \boldsymbol{P}_s^{-1} \boldsymbol{x}}$.
Afterwards, we determine the point's column and row in the current sweep's image frame as $\theta' = \atan2\left(y', x'\right)$ and $\varphi' = \arcsin{\left(z'/r'\right)}$ where $\boldsymbol{x}' = [x', y', z']^T$ and $r' = \|\boldsymbol{x}'\|$.
By repeating this process for each LiDAR point in the previous sweeps, we obtain a mapping between all the previous images, $\{\boldsymbol{I}_S, \boldsymbol{I}_{S+1}, \ldots, \boldsymbol{I}_{-1}\}$, and the current image, $\boldsymbol{I}_0$.
If multiple points from a single sweep map to the same coordinate, we keep the point with the smallest $r'$.
This mapping is used to warp the learned features into the shared image frame defined by the current sweep.

\subsubsection{Multi-Sweep Architecture}
The range images corresponding to the individual sweeps are passed into a CNN where the parameters are shared across all sweeps.
By sharing the weights, the network is forced to extract the same set of features from each sweep.
Afterwards, the extracted features from the previous sweeps are concatenated with the corresponding ego-motion features and passed into another CNN, which we refer to as the transformer network.
Again the weights are shared across the sweeps, and the purpose of this network is to undo the effect of the ego-motion on the extracted features.
Ideally, after this network, the only remaining difference between the sweep features will be due to the motion of the objects and not the motion of the self-driving vehicle.
The resulting feature maps are warped as described above and concatenated with the current sweep's feature map and passed to a U-Net style backbone network~\cite{unet}.

\begin{figure*}[t]
  \begin{subfigure}{.64\textwidth}
    \begin{subfigure}{.57\textwidth}
        \centering
        \includegraphics[height=10cm]{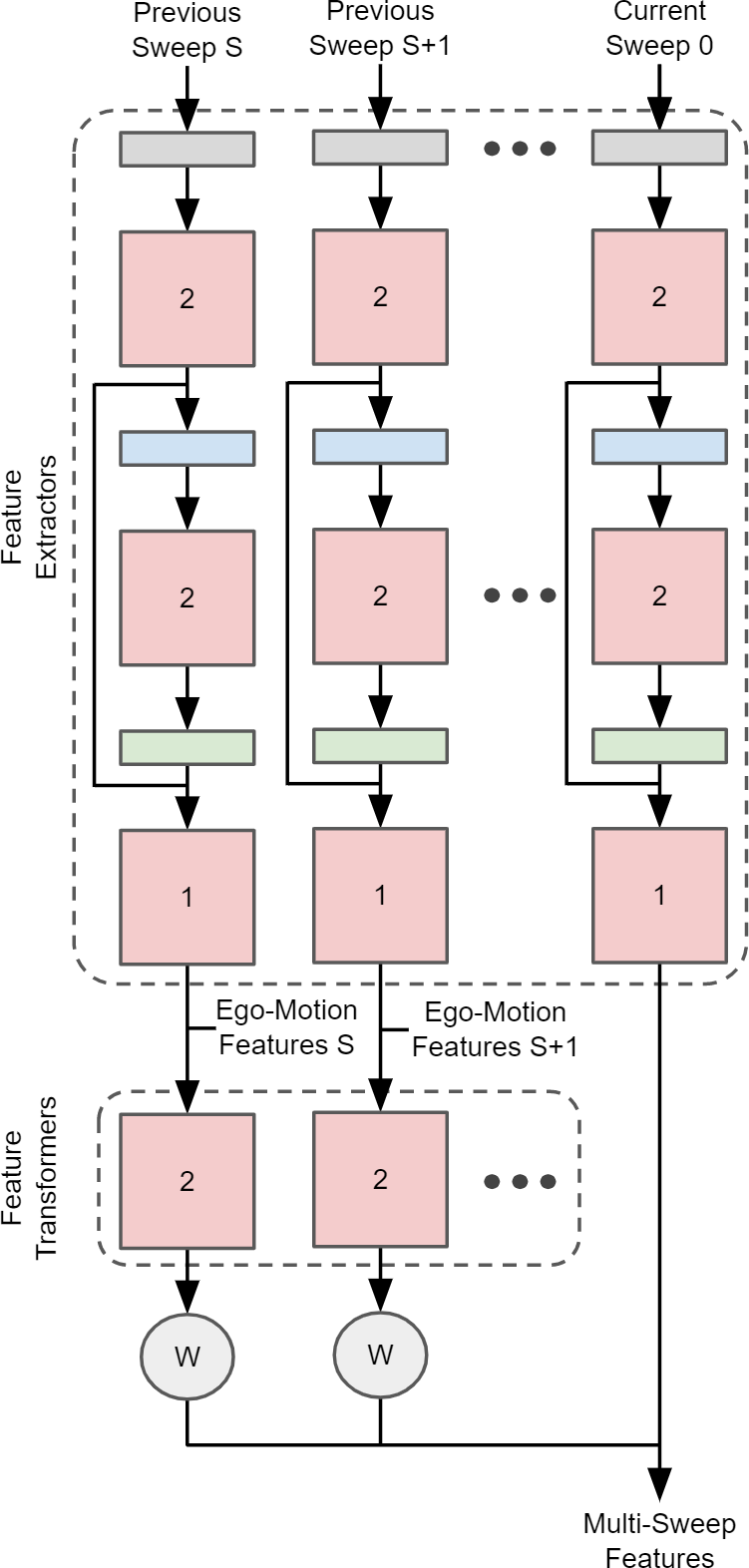}
        \caption{Fusion Network}
    \end{subfigure}
    \begin{subfigure}{.42\textwidth}
        \centering
        \includegraphics[height=8.125cm]{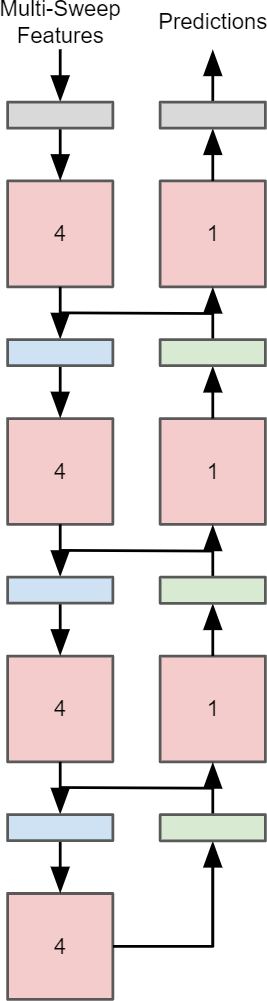}
        \vspace{2cm}
        \caption{Backbone Network}
    \end{subfigure}
  \end{subfigure}
  \begin{subfigure}{.35\textwidth}
    \begin{subfigure}{.98\textwidth}
        \centering
        \vspace{0.125cm}
        \includegraphics[height=2.25cm]{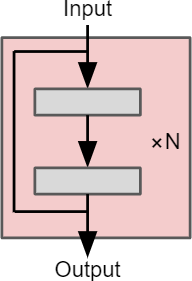}
        \vspace{0.625cm}
        \caption{Residual Block}
    \end{subfigure}
    \begin{subfigure}{.98\textwidth}
        \centering
        \vspace{1cm}
        \includegraphics[height=4.25cm]{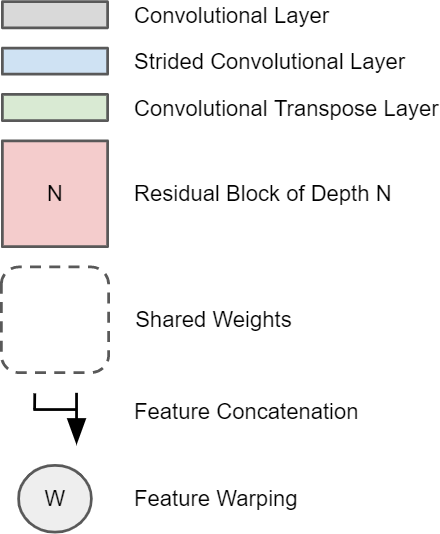}
        \vspace{1cm}
        \caption{Legend}
    \end{subfigure}
  \end{subfigure}
  \caption{The network architecture utilized by our proposed method.}
  \label{fig:network}
  \vspace{-1em}
\end{figure*}

The full network architecture used by our proposed method to extract multi-sweep features and to detect and predict the motion of objects is depicted in Figure \ref{fig:network}.
The network is fully convolutional, and its input is the range view representation of the LiDAR.
Since the height of the range image is significantly smaller than the width, downsampling and upsampling is only performed on the columns of the image, i.e. the number of rows in the image is constant throughout the network.
The number of kernels used for each convolution is $16$ in the multi-sweep fusion network and $64$ in the backbone network.

\subsection{Model Predictions}
\label{sec:predictions}

Our goal is to predict the probability of all object trajectories, $\boldsymbol{\mathcal{T}}$, within the scene given the multi-sweep LiDAR data, $\boldsymbol{\mathcal{X}}$.
We assume all the objects are conditionally independent given the sensor data and the time-steps are only dependent on the previous step; as a result, the probability factorizes into the following:
\begin{equation}
    {p(\boldsymbol{\mathcal{T}}|\boldsymbol{\mathcal{X}}) = \prod_{i=1}^N p(\boldsymbol{\tau}_0^i|\boldsymbol{\mathcal{X}}) \prod_{t=1}^T p(\boldsymbol{\tau}_t^i|\boldsymbol{\tau}_{t-1}^i,\boldsymbol{\mathcal{X}})}
\end{equation}
where $N$ is the number of objects, $T$ is the number of time-steps into the future for which we predict the motion of the objects, and $\boldsymbol{\tau}_t^i$ is the position of the $i$-th object at the $t$-th time-step represented by the corners of its bounding box.
Additionally, we assume the spatial dimensions of the bounding box are conditionally independent and drawn from Laplace distributions; therefore, the probability of the trajectories becomes,
\begin{align}
    p(\boldsymbol{\mathcal{T}}|\boldsymbol{\mathcal{X}}) &= \prod_{t=0}^T \prod_{i=1}^N \prod_{j=1}^D p(\tau_t^{ij}| \nu_t^{ij}, b_t^{ij}) \\
    &= \prod_{t=0}^T \prod_{i=1}^N \prod_{j=1}^D \frac{1}{2b_t^{ij}} \exp\left(-\frac{\left|\tau_t^{ij} - \nu_t^{ij}\right|}{b_t^{ij}}\right)
\end{align}
where $D$ is the dimensionality of the bounding box, and $\nu_t^{ij} \in \mathbb{R}$ and $b_t^{ij} \in \mathbb{R}_+$ are the mean and scale of the Laplace distribution corresponding to the $j$-th dimension of the $i$-th object's bounding box at time $t$.
The mean $\nu_t^{ij}$ and scale $b_t^{ij}$ of the distribution are outputs of our network and are a function of the sensor data as well as the previous predictions when $t \geq 1$.

\subsubsection{Predicting Trajectories}
To predict object trajectories, we need to identify the LiDAR points that lie on objects.
Therefore, we predict a set of class probabilities for each point in the range image at $t=0$.
Assuming a LiDAR point is on an object, the network predicts a probability distribution over bounding box trajectories.
Following \cite{lasernet}, a bounding box is represented by its four corners in the BEV and all predictions are made relative to LiDAR points.
For each point, the network outputs the dimensions of the object's bounding box $(l, w)$, which we assume is constant across time.
Also, the network outputs a set of displacement vectors $\{(d_t^x, d_t^y)\}_{t=0}^T$, a set of rotation angles $\{(\omega_t^x, \omega_t^y) := (\cos 2\omega_t, \sin 2\omega_t)\}_{t=0}^T$, and a set of uncertainties $\{(s_t^x, s_t^y) := (\log b_t^x, \log b_t^y)\}_{t=0}^T$ for all of the time-steps.
At $t=0$, the center and orientation of the bounding box is computed as
\begin{equation}
    \boldsymbol{c}_0 = \left[ x, y \right]^T + \boldsymbol{R}_\theta \left[ d_0^x, d_0^y \right]^T
\end{equation}
and
\begin{equation}
    \phi_0 = \theta + \frac{1}{2}\atan2\left(\omega_0^y, \omega_0^x \right)
\end{equation}
where $\left(x, y \right)$ is the position of the LiDAR point in the BEV and $\theta$ is the azimuth angle of the LiDAR point.
Notice that the network predicts an orientation between $-90^\circ$ and $90^\circ$, since the bounding box is symmetrical.
Furthermore, when $t \geq 1$, the center and orientation is
\begin{equation}
    \boldsymbol{c}_t = \boldsymbol{c}_{t-1} + \boldsymbol{R}_\theta \left[ d_t^x, d_t^y \right]^T 
    \label{eqn:predictions-1}
\end{equation}
and
\begin{equation}
    \phi_t = \phi_{t-1} + \frac{1}{2}\atan2\left(\omega_t^y, \omega_t^x \right)\text{.}
    \label{eqn:predictions-2}
\end{equation}
Like in \cite{lasernet}, the corners of the bounding box are computed from the box parameters.
The along track uncertainty, $b_t^x = \exp s_t^x$, and the cross track uncertainty, $b_t^y = \exp s_t^y$, represent the uncertainties at time $t$ in the corners along the direction of motion and perpendicular to the motion, respectively.
At any particular time-step $t$, the scale of the distribution is assumed to be the same for all four corners.

Lastly, the trajectory predictions from individual LiDAR points are clustered into objects using the approximate mean-shift algorithm proposed in \cite{lasernet}.

\subsection{End-to-end Training}
\label{sec:training}

In the following sections, we describe our loss functions utilized to train our network, and our approach to learning the distribution of object trajectories inspired by curriculum learning.

\subsubsection{Loss Functions}
\label{sec:losses}

Our loss functions consist of the focal loss \cite{retinanet} used to learn the class probabilities for each LiDAR point and the KL divergence \cite{lasernet-kl} used to learn the distribution of trajectories.
The classification loss is defined as
\begin{equation}
  \mathcal{L}_{cls} = \frac{1}{HW}\sum_{i=1}^{HW} \sum_{j=1}^C -[\Tilde{c}_i=j](1-p_{ij})^\gamma \log p_{ij}
\end{equation}
where $p_{ij}$ is the probability of the $j$-th class at the $i$-th pixel, $\Tilde{c}_i$ is the ground-truth class of the $i$-th pixel, $[\cdot]$ is an indicator function, $\gamma$ is the focusing parameter \cite{retinanet}, $C$ is the number of object classes plus a background class, and $W$ and $H$ is the width and height of the range image, respectively.
The regression loss is defined as
\begin{align}
    \begin{split}
        &\mathcal{L}_{reg} = \\
        &\quad\frac{1}{TND} \sum_{t=0}^T \sum_{i=1}^N \sum_{j=1}^D D_{KL}\left(p(\tau_t^{ij}| \Tilde{\nu}_t^{ij}, \Tilde{b}_t^{ij}) \| q(\tau_t^{ij}| \nu_t^{ij}, b_t^{ij})\right)
    \end{split}
\end{align}
where $\nu_t^{ij}$ is the $j$-th dimension of the $i$-th predicted bounding box at the $t$-th time-step, $b_t^{ij}$ is the corresponding predicted uncertainty, and $\Tilde{\nu}_t^{ij}$ and $\Tilde{b}_t^{ij}$ define the ground-truth distribution.
The probability distributions $p(\tau_t^{ij}| \Tilde{\nu}_t^{ij}, \Tilde{b}_t^{ij})$ and $q(\tau_t^{ij}| \nu_t^{ij}, b_t^{ij})$ are assumed to be Laplace distributions; therefore, the KL divergence becomes:
\begin{align}
    \begin{split}
        &D_{KL}\left(p(x| \Tilde{\mu}, \Tilde{b}) \| q(x| \mu, b)\right) = \\
        &\quad\log\frac{b}{\Tilde{b}} + \frac{\Tilde{b} \exp\left(-\frac{\left|\mu - \Tilde{\mu}\right|}{\Tilde{b}} \right) + \left|\mu - \Tilde{\mu}\right|}{b} - 1\text{.}
    \end{split}
\end{align}
Refer to \cite{huber_loss} for a derivation and \cite{lasernet-kl} for an analysis of the loss function.
In order to properly learn the along and cross track uncertainties, the predicted and ground-truth bounding boxes need to be transformed into the appropriate coordinate frame.
Before applying the regression loss, the corners of each bounding box are rotated such that the $x$-axis is aligned to the object's direction of motion and the $y$-axis is perpendicular to the object's motion.
This is accomplished by rotating the predicted and ground-truth corners corresponding to the $i$-th object at time $t$ by $\boldsymbol{R}_{\phi_t^i}^T$ where $\phi_t^i$ is the predicted future orientation of the object.
The total loss utilized by our proposed method is $\mathcal{L}_{total} = \mathcal{L}_{cls} + \lambda \mathcal{L}_{reg}$ where $\lambda$ is used to weight the relative importance of the individual losses.
For all experiments, $\gamma = 2$ and $\lambda = 4$, which we empirically found to perform well.

\subsubsection{Uncertainty Curriculum}
\label{sec:curriculum}

To utilize the KL divergence, the parameters of the ground-truth distribution need to be specified for all objects and time-steps.
The means of the distribution are simply set to the ground-truth bounding box provided by an annotator, but how to set the scales of the distribution remains an open question.
In \cite{lasernet-kl}, heuristics for approximating the uncertainty of the ground-truth were explored for object detection.
In this work, we propose a different approach inspired by curriculum learning \cite{curriculum}.

Our network is trained such that the early predictions affect the future predictions as expressed in Eq. \eqref{eqn:predictions-1} and Eq. \eqref{eqn:predictions-2}.
As a result, early predictions need to be reliable to accurately predict future time-steps.
Based on this observation, at the start of training, we set the uncertainty of the ground-truth for later time-steps to be high while the uncertainty of the early time-steps are set to be low,
\begin{equation}
    \Tilde{b}_t = \alpha b_t^{\text{max}} +  (1 - \alpha) b_t^{\text{min}}
    \label{eq:curriculum}
\end{equation}
where $\alpha \in [0, 1]$, $b_t^{\text{max}} = \frac{t}{T} \eta + \epsilon$, $b_t^{\text{min}} = \epsilon$, and $\Tilde{b}_t$ is shared across all $N$ objects and all $D$ dimensions at time $t$.
As training progresses, the uncertainty of all time-steps is exponentially reduced, ${\alpha = \exp(-\beta k)}$, where $\beta$ controls the rate of decay and $k$ is the training iteration.
As shown in \cite{lasernet-kl}, increasing the ground-truth uncertainty of a particular example reduces the loss for that example as long as the predicted uncertainty does not under-estimate the ground-truth uncertainty.
Therefore, by setting the uncertainty of the earlier time-steps to be lower than the later time-steps, we are placing a larger emphasis on learning the earlier time-steps.
For all of our experiments, $\eta = 100$ cm, $\epsilon = 5$ cm, and $\beta$ is set such that $\alpha \approx 0$ half-way through training.

\begin{table*}[t]
    \centering
    \caption{Detection and Motion Forecasting Performance on ATG4D}
    \begin{tabular}{C{12em} |C{11em} |C{4em} C{4em} C{4em}}
        \hline
        \multirow{2}{*}{Method} & Average Precision (\%) & \multicolumn{3}{c}{$L_2$ Error (cm)} \\
        & 0.7 IoU & 0.0 s & 1.0 s & 3.0 s \\ \hline
        FaF \cite{faf} & 64.1 & 30 & 54 & 180 \\
        IntentNet \cite{intentnet} & 73.9 & 26 & 45 & 146 \\
        NMP \cite{nmp} & 80.5 & 23 & 36 & 114 \\
        SpAGNN \cite{spagnn} & 83.9 & 22 & 33 & \textbf{96} \\
        LaserFlow (ours) & \textbf{84.5} & \textbf{19} & \textbf{31} & 99 \\
        \hline
    \end{tabular}
    \label{tab:atg4d}
\end{table*}
\begin{table*}[t]
    \centering
    \caption{Detection and Motion Forecasting Performance on NuScenes}
    \begin{tabular}{C{12em} |C{11em} |C{4em} C{4em} C{4em}}
        \hline
        \multirow{2}{*}{Method} & Average Precision (\%) & \multicolumn{3}{c}{$L_2$ Error (cm)} \\
        & 0.7 IoU & 0.0 s & 1.0 s & 3.0 s \\ \hline
        SpAGNN \cite{spagnn} & Not Reported & \textbf{22} & 58 & 145 \\
        LaserFlow (ours) & 56.1 & 25 & \textbf{52} & \textbf{143} \\
        \hline
    \end{tabular}
    \label{tab:nuscenes}
\end{table*}

A sensible alternative could be to re-weight the loss for various time-steps and use a constant ground-truth uncertainty.
To show why this is a less attractive option, let us inspect the derivative of the KL divergence with respect to the predicted mean,
\begin{equation}
    \frac{\partial D_{KL}}{\partial \mu} = \frac{\sgn(\mu - \Tilde{\mu})}{b} \left(1 - \exp\left(-\frac{\left|\mu - \Tilde{\mu}\right|}{\Tilde{b}} \right)\right)\text{.}
    \label{eqn:derivative}
\end{equation}
Re-weighting the KL divergence simply re-scales the derivative.
However, by increasing the ground-truth uncertainty, the magnitude of the derivative is only reduced once the ratio between $|\mu - \Tilde{\mu}|$ and $\Tilde{b}$ becomes small, since the exponential term in Eq. \eqref{eqn:derivative} will go to one.
Therefore, at the start of training, the network will be pushed to correct significant errors while ignoring small errors.
By employing a curriculum on the uncertainty, we can ensure the predictions for the early time-steps will be refined before the later time-steps.
In the next section, we demonstrate through experimentation that utilizing an uncertainty curriculum is superior to re-weighting the loss.

\section{Experiments}
\label{sec:experiments}

Our proposed method is evaluated and compared to state-of-the-art end-to-end methods on two autonomous driving datasets: ATG4D \cite{lasernet} and NuScenes \cite{nuscenes}.
The ATG4D dataset contains $1.2 \times 10^6$ sweeps for training where the NuScenes dataset contains $2.7 \times 10^5$ training sweeps.
ATG4D uses a 64-beam LiDAR resulting in a $2048 \times 64$ range image, and NuScenes uses a 32-beam LiDAR producing a range image with a resolution of $1024 \times 32$.
For both datasets, we utilize a total of $5$ input sweeps, and predict $3$ seconds into the future at $0.5$ seconds intervals.
On the NuScenes dataset, we employ data augmentation, i.e. apply random translations, rotations, and dilations to the point cloud and labels, since there are fewer training examples.
To train the network, we use the same optimizer and learning scheduling as \cite{lasernet}.

Following \cite{spagnn}, we measure object detection and motion forecasting performance for vehicles within a region of interest (RoI).\footnote{Our method produces predictions at the full range of the sensor and is not restricted to the RoI.}
For ATG4D, the RoI is $144 \times 80$ meters centered on the self-driving vehicle, and for NuScenes, the RoI is $100 \times 100$ meters.
Detection performance is evaluated using the average precision (AP) metric where an intersection-over-union (IoU) of $0.7$ with a ground-truth label is required for a detection to be considered a true positive.
Furthermore, to evaluate the precision of the motion forecasting, we use the $L_2$ error between the ground-truth's center and the prediction's center at various time-steps.
Since the $L_2$ error is affected by the sensitivity of the detector, \cite{spagnn} uses a fixed recall to evaluate the models.
For ATG4D, the recall point is 80\% at a $0.5$ IoU, and for NuScenes, the recall point is 60\% at a $0.5$ IoU.

Note, our experimentation is focused on motion forecasting, since our approach to object detection follows \cite{lasernet}.
Therefore, we refer the reader to \cite{lasernet} for an in-depth analysis on object detection.

\subsection{Comparisons with the State-of-the-Art}
The results for our proposed approach and existing state-of-the-art methods on ATG4D are shown in Table~\ref{tab:atg4d}.
On this dataset, our method obtains better results compared to the existing methods in terms of AP and $L_2$ error at $0$ and $1$ seconds.
LaserFlow as well as FaF \cite{faf}, IntentNet \cite{intentnet}, and NMP~\cite{nmp} predict trajectories in a single stage; however, SpAGNN \cite{spagnn} uses multiple stages to refine their predictions.
When compared to their single stage version, our method out-performs SpAGNN by $8$ cm at $3$ seconds.
The results on NuScenes are shown in Table~\ref{tab:nuscenes}.
On this dataset, our method out-performs SpAGNN at $1$ and $3$ seconds.

\subsection{Ablation Study}
On ATG4D, we perform an ablation study on our multi-sweep fusion architecture, and the results are listed in Table~\ref{tab:multisweep_ablation}.
As shown in the table, naively applying the early fusion method employed by BEV methods to the RV, i.e. rendering the previous sweeps in the current's sweep coordinate frame, performs poorly at longer time horizons.
Leveraging our proposed fusion method without the transformer network also under-performs at later time-steps showing the benefit of learning this transformation.
Lastly, keeping the ego-motion feature in the global coordinate frame, by removing the rotation from Eq. \ref{eq:egomotion}, has a substantial impact on performance.

In addition, we conduct an ablation on our training procedure, and the results can be seen in Table~\ref{tab:training_ablation}.
We observe that predicting only the mean of the distribution, i.e. removing the uncertainty estimate, significantly degrades the performance in terms of both object detection and motion forecasting.
When predicting the full distribution, we see a reduction in performance at $3$ seconds when the uncertainty curriculum is not utilized ($\alpha = 0$ in Eq. \ref{eq:curriculum}).
Furthermore, applying a curriculum on the weight instead of the uncertainty performs worse at $3$ seconds, which demonstrates the value of our proposed uncertainty curriculum.

\begin{table*}[t]
    \centering
    \caption{Ablation Study on Multi-Sweep Fusion}
    \begin{tabular}{C{12em} |C{11em} |C{4em} C{4em} C{4em}}
        \hline
        \multirow{2}{*}{Experiment} & Average Precision (\%) & \multicolumn{3}{c}{$L_2$ Error (cm)} \\
        & 0.7 IoU & 0.0 s & 1.0 s & 3.0 s \\ \hline
        Early Fusion & 83.7 & 20 & 36 & 115 \\
        No Transformer Network & 83.8 & 20 & 35 & 112  \\
        Global Ego-Motion Feature & 84.0 & 20 & 33 & 107 \\
        \hline
        Proposed Method & \textbf{84.5} & \textbf{19} & \textbf{31} & \textbf{99} \\
        \hline
    \end{tabular}
    \label{tab:multisweep_ablation}
\end{table*}
\begin{table*}[t]
    \centering
    \caption{Ablation Study on the Training Procedure}
    \begin{tabular}{C{12em} |C{11em} |C{4em} C{4em} C{4em}}
        \hline
        \multirow{2}{*}{Experiment} & Average Precision (\%) & \multicolumn{3}{c}{$L_2$ Error (cm)} \\
        & 0.7 IoU & 0.0 s & 1.0 s & 3.0 s \\ \hline
        No Uncertainty & 82.7 & 21 & 35 & 111 \\
        No Learning Curriculum & 83.2 & \textbf{19} & 32 & 104 \\
        Weight-based Curriculum & 84.0 & 20 & 35 & 117 \\
        \hline
        Proposed Method & \textbf{84.5} & \textbf{19} & \textbf{31} & \textbf{99} \\
        \hline
    \end{tabular}
    \label{tab:training_ablation}
\end{table*}

\begin{figure*}[t]
  \begin{subfigure}{.48\textwidth}
    \centering
    \includegraphics[width=.8\linewidth]{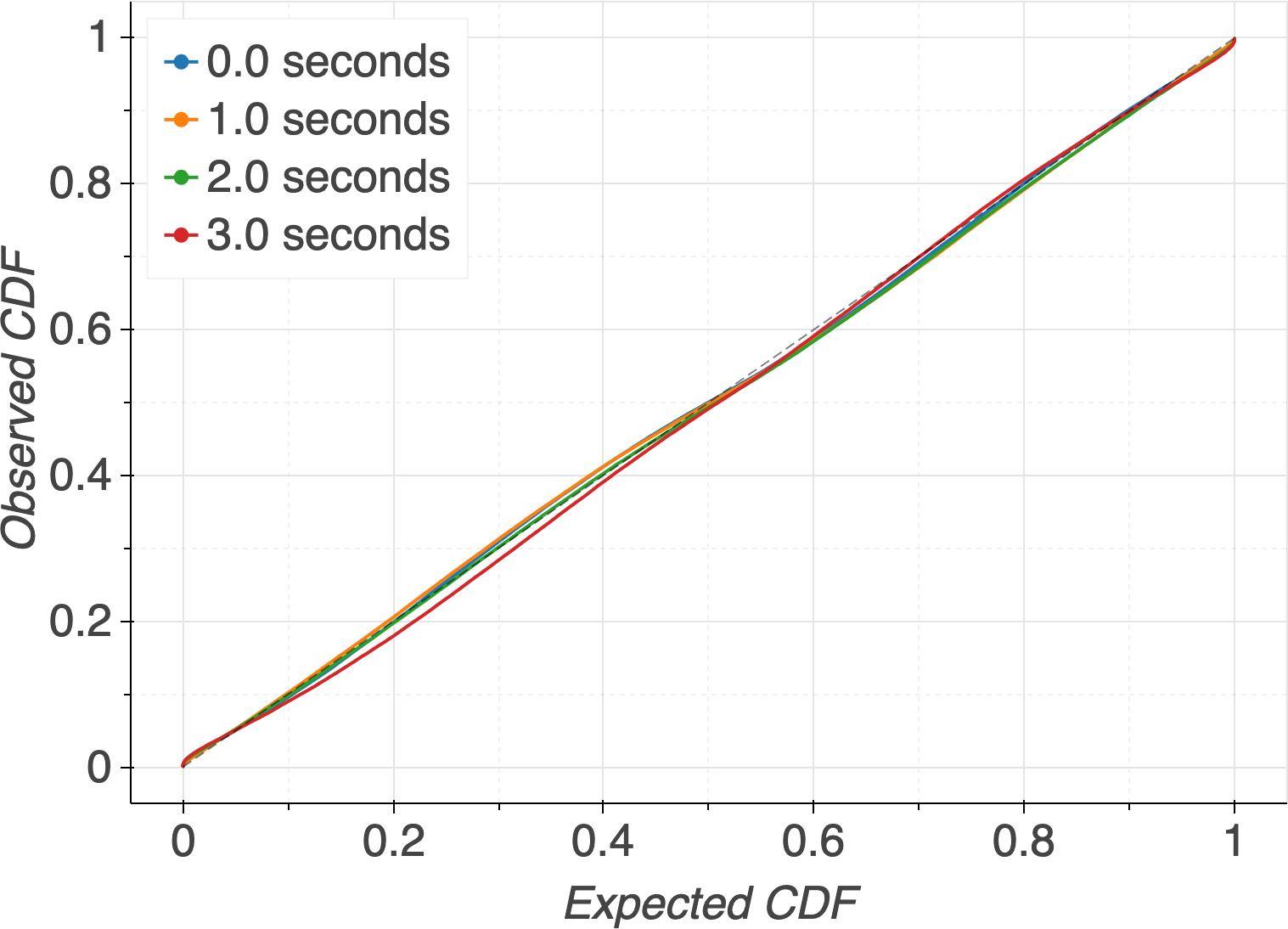}
    \caption{Along-Track Calibration}
  \end{subfigure}
  \begin{subfigure}{.48\textwidth}
    \centering
    \includegraphics[width=.8\linewidth]{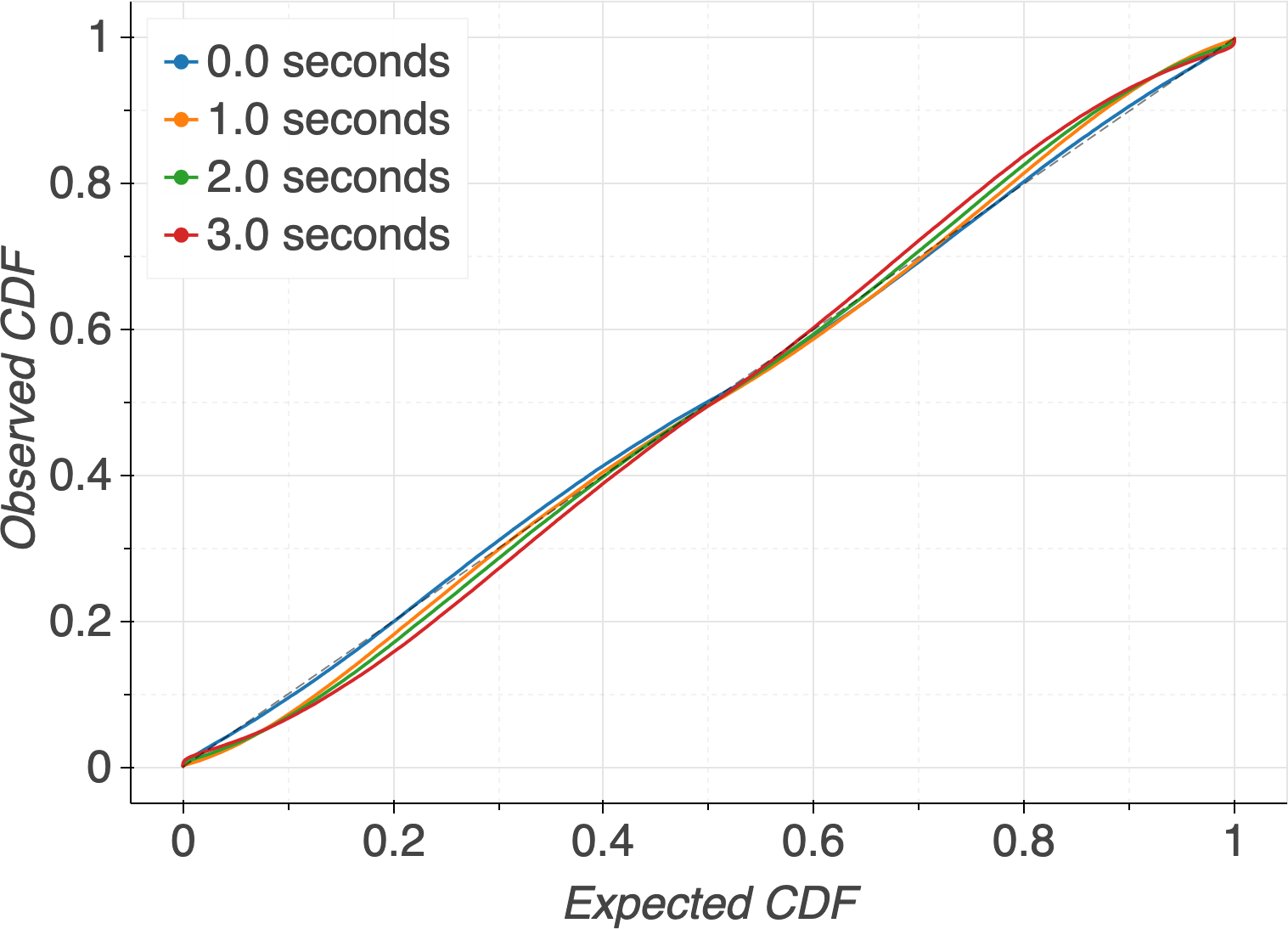}
    \caption{Cross-Track Calibration}
  \end{subfigure}
  \caption{Calibration plots showing the reliability of the predicted distributions at different time-steps. A well-calibrated distribution will follow the dashed line.}
  \label{fig:calibration}
\end{figure*}

\subsection{Uncertainty Evaluation}
The calibration of the predicted distributions on ATG4D is shown in Figure~\ref{fig:calibration}.
To evaluate the predicted probability distribution, we compare the expected cumulative distribution function (CDF) to the observed CDF.
An in-depth explanation on how the calibration plots are generated is available in \cite{lasernet-kl}.
From the figure, we observe that our proposed method can accurately learn the along and cross track distribution of the bounding box corners for all time-steps.

\subsection{Runtime Performance}
The total runtime of our proposed method on ATG4D is $60$ ms.
IntentNet~\cite{intentnet}, NMP~\cite{nmp}, and SpAGNN~\cite{spagnn} operate at $100$ ms or greater.
FaF~\cite{faf} has a runtime of $30$ ms; however, their BEV method operates only on the $144 \times 80$ meter RoI.
Our RV method operates at the full range of the LiDAR which is a circle with diameter of $240$ meters, and its runtime is independent of the range of the sensor.
Therefore, our method runs on an area approximately $4$x the size in only $2$x the time.

For the purposes of autonomous driving, it is critical to consider the entire scene without disregarding any measurements provided by the sensor.
Imagine making an unprotected right-hand turn onto a $45$ mph street. If LiDAR points beyond $40$ m to the left are discarded, then the autonomous vehicle has less than $2$ seconds to react to a vehicle traveling at the speed limit. By operating at the full range of the sensor, the vehicle has $3$x the amount of time to react.

\subsection{Multi-Class Object Detection and Motion Forecasting}
Our method utilizes the range view representation of the LiDAR; therefore, it operates at the full range of the sensor without voxelization.
As a result, it is suitable for detecting both large and small objects, e.g. vehicles as well as pedestrians and bicycles, with a single network.
In Table \ref{tab:multiclass}, we list the object detection and motion forecasting performance of our multi-class model on ATG4D.
In this case, the region-of-interest (RoI) is a circle with a diameter of $240$ meters centered on the self-driving vehicle, which encapsulates the entire range of the LiDAR.
To evaluate object detection, we require pedestrian and bicycle detections to have an intersection-over-union (IoU) with a ground-truth of at least $0.5$ to be considered a true positive.
To evaluate motion forecasting, we use a recall point of 80\% at a $0.5$ IoU for pedestrians and a recall point of 60\% at a $0.5$ IoU for bicycles.

From Table \ref{tab:multiclass}, we observe that adding multiple classes does not significantly impact the model's ability to detect and predict the motion of vehicles.
Of the three classes, bicycles are the hardest to detect due to their rarity in the training set.
However, the $L_2$ error for bicycles is the lowest at longer time horizons because their speed is lower than vehicles and their movement is more consistent than pedestrians.
None of the existing work detects and predicts motion for multi-classes or operates at the full range of the sensor so we cannot show performance comparisons, but these results show the ability of our approach to extend to multiple classes.

\begin{table*}
    \centering
    \caption{Full Sensor Range Multi-Class Detection and Motion Forecasting Performance on ATG4D}
    \begin{tabular}{c|c|ccc|c|ccc|c|ccc}
        \hline
        \multirow{3}{*}{Model} & \multicolumn{4}{c|}{Vehicle} & \multicolumn{4}{c|}{Bicycle} & \multicolumn{4}{c}{Pedestrian} \\ \cline{2-13}
        & AP (\%) & \multicolumn{3}{c|}{$L_2$ (cm)} & AP (\%) & \multicolumn{3}{c|}{$L_2$ (cm)} & AP (\%) & \multicolumn{3}{c}{$L_2$ (cm)} \\
        & 0.7 IoU & 0.0 s & 1.0 s & 3.0 s & 0.5 IoU & 0.0 s & 1.0 s & 3.0 s & 0.5 IoU & 0.0 s & 1.0 s & 3.0 s \\ \hline
        Vehicle Only & 74.6 & 24 & 38 & 111 & - & - & - & - & - & - & - & - \\
        Multi-Class & 74.5 & 24 & 38 & 112 & 59.1 & 15 & 22 & 50 & 74.8 & 12 & 27 & 72 \\
        \hline
    \end{tabular}
    \label{tab:multiclass}
    \vspace{-1em}
\end{table*}

\section{Conclusion}
\label{sec:conclusion}

We proposed a novel method for 3D object detection and motion forecasting from the native range view representation of the LiDAR.
Our approach is both efficient and competitive with existing state-of-the-art methods that rely on the bird's eye view representation.
Although we focused on demonstrating the effectiveness of the range view at these tasks, we believe our multi-sweep fusion and uncertainty curriculum could be successfully incorporated into a bird's eye view or a hybrid view system to exploit the advantages of both representations.

\bibliographystyle{IEEEtran}
\bibliography{bibtex}

\end{document}